# SCAN: SITE-AWARE COLLABORATIVE PRIVACY-PRESERVING ADAPTATION OF VISION-LANGUAGE MODELS FOR NEUROLOGICAL DISEASE CLASSIFICATION

*Nikhil J. Dhinagar*[1], Rishab Pendam[1], Harmony S. Cen[1], Sophia I. Thomopoulos[1], John P. John[2], Mahir H. Khan[3], Sook-Lei Liew[3], Erik Kaestner[4], Carrie McDonald[4], Paul M. Thompson[1]*

[1] Imaging Genetics Center, Mark & Mary Stevens Neuroimaging & Informatics Institute, Keck School of Medicine, University of Southern California, Los Angeles, CA, USA
[2] Multimodal Brain Image Analysis Laboratory, ADBS Neuroimaging Centre, National Institute of Mental Health and Neurosciences (NIMHANS), Bangalore, India
[3] Neuroscience Graduate Program, Mark & Mary Stevens Neuroimaging & Informatics Institute, Chan Division of Occupational Science & Occupational Therapy, Biomedical Engineering, University of Southern California, Los Angeles, CA, USA
[4] Department of Radiation Medicine and Applied Sciences, University of California, San Diego, San Diego, CA, USA

## ABSTRACT

Worldwide neuroimaging data could strengthen diagnostic foundation models, but privacy rules often prevent centralizing brain MRI scans. We present SCAN, a site-aware collaborative adaptation framework for Neurological Disease Classification that lets new sites adapt a 3D medical vision-language model (VLM) without retraining or sharing patient scans. Each site keeps exact class prototypes for local prediction and can share privacy-protected class summaries: norm-bounded sums and counts perturbed once with calibrated noise. Using 7,641 scans from 67 sites across 12 countries, covering Alzheimer's disease (AD), stroke, epilepsy and controls, we assess local adaptation, cross-site benefits, and effects on existing sites. At seven unseen sites, the frozen VLM's zero-shot AUCs were 0.833-0.945 for AD and stroke. Training-free local prototypes improved epilepsy AUC at three sites from 0.598/0.625/0.508 to 0.684/0.729/0.762 and AD AUC at one site from 0.833 to 0.863, nearly matching a classifier trained at each site (mean AUC 0.757 vs. 0.759). With differential privacy ($\varepsilon = 4$), new sites contributed without materially reducing accuracy at existing sites (AUC change $\leq$0.2%), although gains from sharing were limited, and membership attacks performed near chance. SCAN offers hospitals a worldwide framework to adapt and share brain MRI VLMs without exchanging scans or retraining.

## 1. INTRODUCTION

Deep models for brain MRI often generalize poorly to sites they were not trained on, because scanners, acquisition protocols and patient populations differ. Consortia such as ENIGMA [1] show the value of combining data across many institutions, yet pooling images is often restricted by data use agreements and institutional preference to keep scans on site. Federated learning [2-4] avoids centralizing raw images, but requires repeated rounds of model updates, training infrastructure and local IT co-operation at every site, and shared updates can still leak patient information [5]. Approaches that let institutions contribute to shared models without transferring raw scans could therefore improve diagnostic models while limiting these privacy risks.

Foundation models for medical image analysis have considerable potential across clinical sites, but adapting them to a new hospital typically requires fine-tuning, which is computationally demanding and needs sufficient local training data. Alternatively, a pretrained vision-language model (VLM) [6] can be adapted without changing its weights, using class prototypes (average feature representations of labeled scans) to classify new images by their similarity to these prototypes [7, 8].

Class prototypes are compact summaries of labeled MRI scans that can be shared between sites, as in federated prototype learning [9]. Yet because they summarize features of individual patients, shared prototypes may reveal whether a particular patient contributed, especially in small cohorts [10, 11]. Differential privacy (DP) offers a formal way to limit this risk [12]. Conventional privacy-preserving training methods, such as DP-SGD [13], repeatedly add noise to model updates, potentially reducing accuracy, particularly with small medical imaging datasets [14]. In contrast, the proposed site-aware collaborative adaptation framework for Neurological Disease Classification (SCAN) uses each site's original prototypes for local predictions, and adds calibrated noise to the summary statistics once, before sharing them, without retraining the VLM or repeatedly exchanging model updates.

Evaluations of collaborative learning methods often report pooled accuracy without separating the benefits of local adaptation, gains from other sites, and effects on existing participants. We address these questions through three contributions:

(1) SCAN, a framework for adapting a frozen 3D VLM using local site-class prototypes, with a single privacy-protected release using the analytic Gaussian mechanism [15]; (2) an evaluation framework measuring local adaptation, benefits from other sites, and effects on existing cohorts, including a trained classifier baseline and a test of whether individual patients can be identified as contributors; and (3) experiments across seven unseen sites and three neurological diseases, showing that training-free adaptation provides the greatest gains where the VLM performs poorly, while privacy-protected contributions largely preserve accuracy at existing sites.

## 2. METHODS

**2.1 Model Architecture.** A vision-language model (VLM) based on the Contrastive Language-Image Pretraining (CLIP) architecture [6, 16] was trained on 4,773 T1-weighted MRI scans from 60 source sites, after which its weights remain frozen. Training used a symmetric contrastive (InfoNCE) objective, pairing each scan with a caption of the form '{age}-year-old {sex} subject diagnosed as {diagnosis}' and variations, for 20 epochs with optimal hyperparameters. Our primary experimental model uses a 3D DenseNet121 convolutional neural network (11M parameters) as the vision encoder and ClinicalBERT (110M parameters) as the text encoder [16, 17]. Its image encoder $\varphi$ converts each T1-weighted MRI scan $x$ into a 768-dimensional feature vector, normalized as $f = \varphi(x)/\|\varphi(x)\| \in \mathbb{R}^d$. Zero-shot class probabilities are $p(x) = \mathrm{softmax}(s(x)/\tau)$, where $s(x)$ is the cosine similarity between $f$ and the text embedding of class $c \in \{\mathrm{CN}, \mathrm{AD}, \mathrm{stroke}, \mathrm{epilepsy}\}$ and $1/\tau = 5.68$ is the model's learned calibration temperature scale [18]. A new site $k$ holds a small labeled support set (20% of dataset) and wants to classify its remaining scans (evaluation set). Fig. 1 summarizes the protocol.

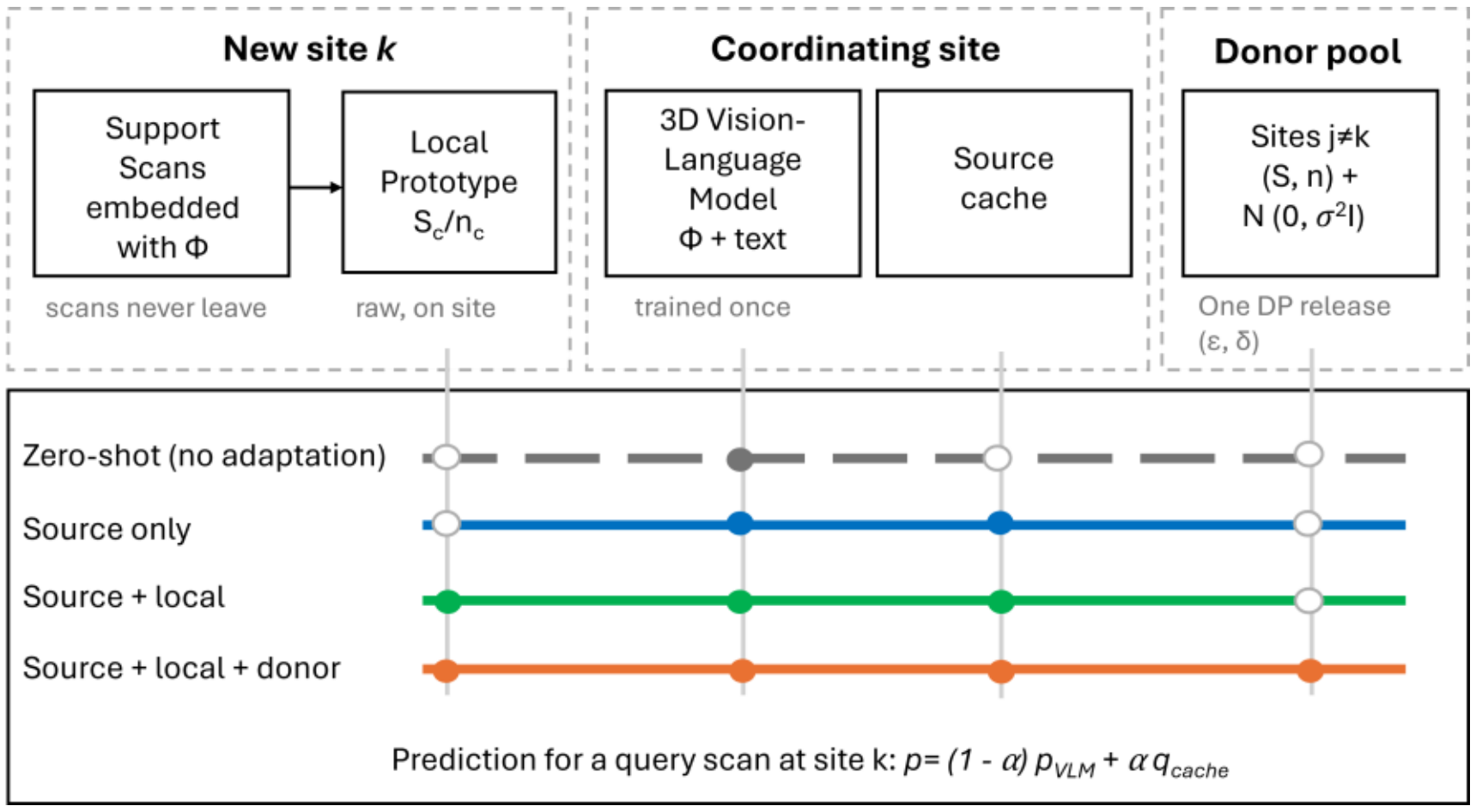

**Figure 1.** Overview of SCAN. A coordinating site keeps the frozen vision-language model (VLM) and prototypes from the original training sites (source cache). Each new site computes class prototypes from its own MRI scans without sharing images, and can share prototypes protected by Gaussian noise on class sums and counts ((ε, δ)-differential privacy) that are reused without retraining the VLM. Rows: zero-shot VLM and three adaptation strategies; filled dots mark the components used.

**2.2 Site-class prototypes** For site $k$ and diagnostic class $c$, SCAN summarizes the labeled MRI embeddings using their sum and count:

$$S_{k,c} = \sum\nolimits_{i:s_i=k,y_i=c} f_i, \quad n_{k,c} = \left|\left\{i: s_i = k, y_i = c\right\}\right| \qquad (1)$$

The site–class prototype is the mean embedding. Classes with fewer than $m$ scans are excluded. SCAN uses three prototype sets (caches): (1) a source cache built by the coordinating site from the original training sites ($m = 10$); (2) a local cache built from each new site's labeled support scans ($m = 2$), retained at that site; and (3) a donor cache of privacy-protected prototypes contributed by other new sites ($m = 2$).

**2.3 Protected statistical contribution.** To protect patient privacy when sharing prototypes, SCAN adds calibrated Gaussian noise to each site's class-wise embedding sums and counts. Under differential privacy, neighboring datasets differ by the addition or removal of one scan. Since each embedding has an L2 norm of at most one, adding or removing a scan changes its class sum by at most one and its count by one, giving a joint L2 sensitivity of $\sqrt{2}$. Each contributing site releases its statistics once:

$$\left(\hat{S}_c, \hat{n}_c\right) = \left(S_{k,c}, n_{k,c}\right) + \mathcal{N}\left(0, \sigma^2 I\right) \qquad (2)$$

The noise scale σ is calibrated using the analytic Gaussian mechanism [15], implemented in *diffprivlib* [19], to satisfy (ε, δ)-DP; for $\delta = 10^{-5}$, $\sigma = 1.53$ at $\varepsilon = 4$ and $\sigma = 5.28$ at $\varepsilon = 1$.

The receiving site retains a class $c$ only if its noisy count satisfies $\hat{n}_c \geq \max(m, 3\sigma)$, where $m$ is the minimum count. This excludes classes with too few scans for reliable prototypes. For each retained class, the prototype is computed by dividing the noisy embedding sum by the noisy count, $\hat{\mu}_c = \hat{S}_c/\hat{n}_c$, and clipping its L2 norm to at most one. Because these steps use only privacy-protected statistics and public parameters, they preserve the differential privacy guarantee through post-processing [12].

Differential privacy protects only new sites' contributions, not the VLM and source cache held by the trusted coordinating site. Each new site retains its exact local prototypes for diagnosis but can also use privacy-protected prototypes shared by other sites; shared prototypes can be added to or removed from the cache without retraining the VLM.

**2.4 Prediction with prototype cache.** For each evaluation scan, SCAN compares its normalized MRI embedding with the cached prototypes using dot-product similarity. Given a cache with prototypes $\mu_j$, let $J_c$ denote the prototypes belonging to diagnostic class $c$. Similarities are scaled by a retrieval temperature $T = 0.1$ and combined within each class using log-mean-exp pooling, which emphasizes the most similar prototypes while accounting for differing numbers of prototypes per class. A softmax converts the pooled class scores into probabilities over $c$. These are used to adjust the VLM's predictions for classes represented in the cache, while leaving probabilities for other classes unchanged.

**Table 1.** Datasets: ADNI (Alzheimer's Disease Neuroimaging Initiative), OASIS (Open Access Series of Imaging Studies) and NIMHANS (National Institute of Mental Health and Neurosciences) for AD; ENIGMA (Enhancing Neuroimaging Genetics through Meta-Analysis) Stroke Recovery and Epilepsy working groups. Countries: ADNI, OASIS3: USA; NIMHANS: India; ENIGMA-Stroke: USA, Italy, Canada, Brazil, Norway, UK, Germany, Spain, Australia, China; ENIGMA-Epilepsy: USA, Italy, Canada, Brazil, Mexico, UK, Germany, Australia. *Age missing for 1 epilepsy site, sex missing for 18 epilepsy sites and 1 stroke site.

| Dataset | Sites | Scans (subj.) | Sex M/F | Age, mean (SD), range | CN / disease |
|---|---|---|---|---|---|
| ADNI | 1 | 3,236 (1,188) | 1,728/ 1,508 | 75.7 (6.9) 55-94 | 1,864 / 1,372 AD |
| OASIS3 | 1 | 600 (600) | 259/ 341 | 70.6 (9.4) 43-97 | 417 / 183 AD |
| NIMHANS | 1 | 226 (226) | 125/ 101 | 60.9 (13.3) 50-86 | 128 / 98 AD |
| ENIGMA-Stroke | 47 | 1,579 (1,579) | 962/ 616 | 61.1 (13.1) 17-93 | 160 / 1,419 stroke |
| ENIGMA-Epilepsy | 17 | 2,000 (2,000) | 842/ 1,140 | 36.9 (12.3) 8-81 | 1,058 / 942 TLE |

**Table 2.** AUC on held-out evaluation scans of each unseen site. Sup./Eval.: control/disease counts. Prototype columns blend the zero-shot VLM with the indicated cache; Probe: logistic regression on the site's support embeddings. Brackets: change vs. 0-shot in AUC percentage points; *95% CI of the change excludes 0 (paired bootstrap; uncorrected).

| Site | Dx | Sup. | Eval. | 0-shot | Local proto | Source cache | Source cache + local proto | Probe |
|---|---|---|---|---|---|---|---|---|
| 62 | AD | 83/37 | 334/146 | 0.833 | 0.863 (+3.0%)* | 0.856 (+2.3%)* | 0.865 (+3.2%)* | 0.858 (+2.5%) |
| 63 | stroke | 5/12 | 20/46 | 0.891 | 0.891 (+0.0%) | 0.899 (+0.8%) | 0.900 (+0.9%) | 0.889 (−0.2%) |
| 64 | epilepsy | 7/7 | 29/27 | 0.570 | 0.427 (−14.3%) | 0.531 (−3.8%) | 0.525 (−4.5%) | 0.421 (−14.8%) |
| 68 | epilepsy | 15/6 | 61/24 | 0.598 | 0.684 (+8.6%) | 0.610 (+1.2%) | 0.618 (+2.0%) | 0.676 (+7.8%) |
| 72 | epilepsy | 52/28 | 210/112 | 0.625 | 0.729 (+10.4%)* | 0.603 (−2.1%) | 0.659 (+3.4%) | 0.733 (+10.9%)* |
| 73 | epilepsy | 12/9 | 49/37 | 0.508 | 0.762 (+25.4%)* | 0.561 (+5.3%) | 0.624 (+11.6%)* | 0.789 (+28.1%)* |
| 75 | AD | 26/20 | 102/78 | 0.945 | 0.945 (+0.0%) | 0.947 (+0.2%) | 0.934 (−1.1%) | 0.947 (+0.2%) |
| Mean | | | | 0.710 | 0.757 (+4.7%)* | 0.715 (+0.6%) | 0.732 (+2.2%) | 0.759 (+4.9%)* |

The resulting cache-adjusted distribution, $q_{\text{cache}}$, is blended with the calibrated VLM distribution, $p_{\text{VLM}}$:

$$\hat{p} = (1-\alpha)p_{\text{VLM}} + \alpha q_{\text{cache}} \quad (3)$$

where $\alpha = 0.5$ gives equal weight to both predictions.

For each disease $d$, classification against cognitively normal controls uses the score $\log\hat{p}_d - \log\hat{p}_{\text{CN}}$. The adapter is training-free [7, 8], while retaining multiple site-specific prototypes per class and combining predictions in probability space.

## 3. EXPERIMENTAL SECTION

**3.1 Datasets.** We analyzed 7,641 T1-weighted brain MRI scans from 67 sites across 12 countries, covering AD, stroke, epilepsy, and healthy controls as shown in Table 1. The training set included 4,773 scans: 2,577 from ADNI, 1,077 from ENIGMA-Stroke, and 1,119 from ENIGMA-Epilepsy. Validation used 488 scans (ADNI: 302; ENIGMA-Stroke: 120; ENIGMA-Epilepsy: 66), and testing used 786 scans (ADNI: 357; ENIGMA-Stroke: 299; ENIGMA-Epilepsy: 130). To assess generalization to previously unseen sites, we evaluated a separate zero-shot test set of 1,594 scans: 600 from OASIS, 226 from NIMHANS, 83 from ENIGMA-Stroke, and 685 from ENIGMA-Epilepsy. In ADNI, unique subjects are restricted to a specific fold to avoid data leakage given the longitudinal data. We note that all our test splits have only a single scan per subject.

**3.2 Image Preprocessing.** Following prior work, all 3D T1-weighted brain MRI scans were processed with a standard neuroimaging pipeline: skull-stripping, N4 bias field correction [20], 9-degree-of-freedom linear registration to a template [21], and isotropic resampling to 2 mm voxels. Before training, images were z-score normalized and resized to 128×256×256 voxels, matching the default Med3DVLM configuration [16].

**3.3 Protocol.** At each unseen site, 20% of each class was randomly sampled as the support set and the rest held out for evaluation (Table 2); all methods are scored on the same held-out scans by disease-versus-control area under the ROC curve (AUC). We fix $\alpha = 0.5$, $T = 0.1$, $m = 10$ (source) and $m = 2$ (local and donor), $\delta = 10^{-5}$ and $\varepsilon \in \{4, 1\}$; DP results are averaged over 20 noise draws. (1) Local site adaptation: we compare the VLM alone (0-shot), local prototypes, the source cache, and source plus local prototypes; as a trained baseline we fit an L2-regularized, class-balanced logistic regression on the support embeddings ($C = 1$). 95% confidence intervals for the change in AUC relative to 0-shot were estimated by stratified bootstrap over each site's evaluation scans (2,000 resamples), paired across methods. (2) Benefit from other sites: starting from source plus local, we added the released prototypes of the other new sites, grouped by donor diagnostic class, unprotected or at $\varepsilon \in \{4, 1\}$. (3) Known-cohort impact: on the known-site test split, we compared each disease's cases against controls from the same sites, using the source cache with and without every new site's release. (4) Privacy audit: we ran a distance-to-prototype membership inference attack [10, 11, 22]. Members are a site's support scans of class $c$, non-members are held-out scans of the same site and class, and each candidate is scored by its negative squared distance; withheld classes receive AUC 0.5. DP bounds on attack AUC and TPR were computed from the exact privacy profile of the analytically calibrated Gaussian mechanism [15, 23].

## 4. RESULTS

**Local adaptation.** The zero-shot VLM performed better for AD and stroke (0.833-0.945) than epilepsy (0.508-0.625; Table 2). Local prototypes increased AUC at three epilepsy sites (68, 72, 73) from 0.598, 0.625, and 0.508 to 0.684, 0.729, and 0.762 (+8.6% to +25.4%), and at AD site 62 by 3.0%. Gains were significant at sites 62, 72 and 73 and on average across all sites (+4.7%, 95% CI 1.6-8.1; Fig. 2). At sites 63 and 75, where the VLM already exceeded 0.89, AUC was unchanged (0.0%; 95% CIs -3.5 to +3.9 and -1.2 to +1.4). At epilepsy site 64, with only seven support scans per class, AUC fell from 0.570 to 0.427, although not significantly (95% CI -30.4 to +2.0); a trained linear probe also performed poorly (0.421), suggesting insufficient local data. Overall, training-free local prototypes matched the trained linear probe (mean AUC 0.757 vs. 0.759).

**Donor benefit.** Relative to each site's source and local caches (Table 3), unprotected same-disease donors changed AUC by -5.6

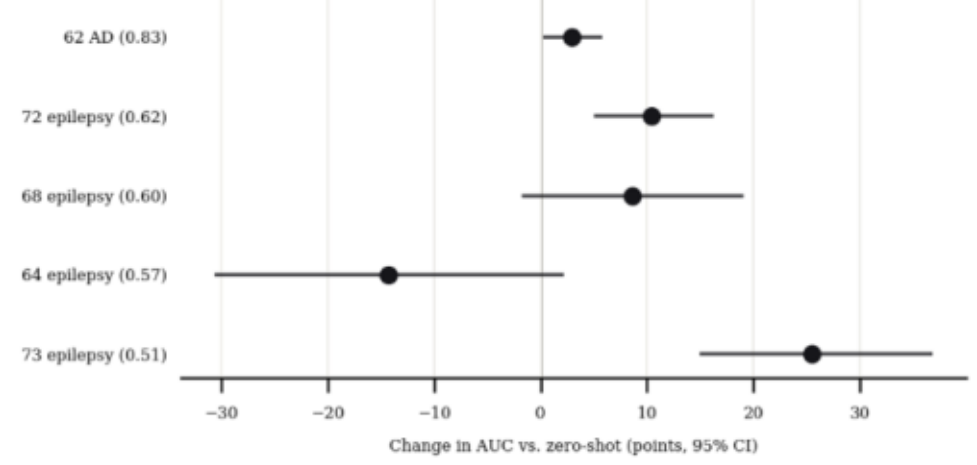

**Figure 2.** Change in AUC (percentage points; 95% paired bootstrap CI) from local prototypes vs. the zero-shot VLM. Sites are ordered by zero-shot AUC (in parentheses), strongest at top; CIs not crossing zero indicate a significant change.

**Table 3.** AUC at recipient sites when same-disease donor prototypes are added to the source cache + local prototypes, unprotected or with DP (mean ± SD over 20 noise draws). Last row: min/max change with cross-disease donors; site 63 has no same-disease donor. Brackets: change vs. source cache + local proto (AUC percentage points).

| Recipient | Dx | Donors | Source cache + local proto | + donor proto (unprotected) | + donor proto (protected ε = 4) | + donor proto (protected ε = 1) |
|---|---|---|---|---|---|---|
| 62 | AD | 75 | 0.865 | 0.862 (−0.3%) | 0.865±0.000 (+0.0%) | 0.864±0.001 (−0.1%) |
| 64 | epilepsy | 68, 72, 73 | 0.525 | 0.469 (−5.6%) | 0.519±0.001 (−0.6%) | 0.525±0.001 (+0.0%) |
| 68 | epilepsy | 64, 72, 73 | 0.618 | 0.589 (−2.9%) | 0.612±0.001 (−0.7%) | 0.618±0.000 (+0.0%) |
| 72 | epilepsy | 64, 68, 73 | 0.659 | 0.665 (+0.7%) | 0.660±0.001 (+0.2%) | 0.658±0.002 (−0.1%) |
| 73 | epilepsy | 64, 68, 72 | 0.624 | 0.644 (+2.0%) | 0.626±0.001 (+0.2%) | 0.623±0.000 (+0.0%) |
| 75 | AD | 62 | 0.934 | 0.935 (+0.1%) | 0.935±0.000 (+0.2%) | 0.936±0.000 (+0.2%) |
| Cross-disease donors (min/max change) | | | | (−0.9%/+0.6%) | (−0.6%/+0.1%) | (−0.4%/+0.2%) |

to +2.0 percentage points and cross-disease donors by -0.9 to +0.6 points. With differential privacy, these effects were smaller: -0.7 to +0.2 points at ε = 4 and -0.1 to +0.2 at ε = 1. Differences may reflect donor numbers and support set sizes rather than diagnosis alone. Adding donors without the source cache caused larger declines (e.g., -19.7 points unprotected and -6.4 at ε = 4 at site 68), suggesting that source prototypes stabilize predictions when incorporating information from other sites.

**Known-cohort impact.** On the known-site test split (same-site controls/cases: AD 170/187; stroke 25/272; epilepsy 52/67), the zero-shot VLM achieved AUCs of 0.925 (AD), 0.869 (stroke) and 0.791 (epilepsy). Adding the source cache reduced AUC by 0.1-0.3 percentage points. Relative to the source cache alone, adding all new sites' prototypes changed AUC by -0.3 to +0.5 points without privacy protection and by at most 0.2 points with differential privacy (ε = 4 or 1; SD ≤ 0.001 over 20 noise draws). Thus, incorporating new sites largely preserved diagnostic performance at existing sites, although it did not improve accuracy.

**Privacy audit.** Across 14 site-class groups, the membership attack [22] performed near chance on unprotected prototypes (mean AUC 0.524; maximum 0.704; TPR 0.104 at 10% FPR, versus 0.10 by chance; Table 4). With differential privacy, mean AUC fell to 0.520 (ε = 4) and 0.510 (ε = 1), below theoretical bounds of 0.743 and 0.575. At ε = 1, approximately half the prototypes were withheld, and TPR was 0.121, below its bound of 0.155.

## 5. DISCUSSION

**Key Findings.** SCAN addresses two challenges in collaborative neuroimaging: adapting pretrained VLMs to new clinical sites without retraining, and sharing information while protecting patient privacy and preserving diagnostic accuracy. Simple site-specific prototypes greatly improved classification where the original VLM generalized poorly (+8.6 to +25.4 points at three epilepsy sites; +3.0 at an AD site) and, without training, matched a locally trained linear classifier (mean AUC 0.757 vs. 0.759).

SCAN also shows that privacy-preserving collaboration is feasible across hospitals. New sites contributed protected prototypes without retraining the VLM, while maintaining diagnostic accuracy at existing sites (AUC changes ≤0.2 percentage points at ε = 4).

**Table 4.** Membership-inference test across 14 site-class groups. Kept: fraction of noise draws yielding a released prototype. Mean AUC assigns 0.5 to withheld prototypes; maximum AUC and true positive rate (TPR) use released prototypes only. Bound: theoretical maximum AUC/TPR under (ε, δ)-DP.

| Release | Kept | AUC mean (max) | TPR @ 10% FPR | Bound |
|---|---|---|---|---|
| Raw | 1 | 0.524 (0.704) | 0.104 | 1.00/1.00 |
| ε = 4 | 0.96 | 0.520 (0.655) | 0.138 | 0.74/0.36 |
| ε = 1 | 0.49 | 0.510 (0.589) | 0.121 | 0.58/0.16 |

Although gains at recipient sites were modest (≤0.7 points), the source cache helped limit negative transfer. These findings establish a practical foundation for expanding shared neurological disease models across international consortia.

**Comparison with federated learning.** Unlike federated learning, which requires repeated rounds of training and model exchange at every site, SCAN needs only a forward pass and one protected release per site.

**Protection against image reconstruction.** SCAN compresses millions of MRI voxels into 768-dimensional embeddings, averages them across patients, and adds calibrated noise. These steps make reconstruction of individual scans difficult, while DP formally limits additional information leakage from each contributed scan [24, 25].

**Comparison with synthetic data.** Synthetic images generated without DP offer no formal privacy guarantee [26], and diffusion models can memorize training scans, including brain MRI [27, 28]; SCAN shares only protected class summaries with an explicit (ε, δ)-DP guarantee.

Although adding prototypes generally preserved existing-site accuracy, each site should validate adaptation on held-out scans, particularly with small support sets (site 64), and further validation is needed before clinical deployment.

**Limitations and Future Work.** We evaluated a single VLM (3D DenseNet121 with ClinicalBERT) and a single support/evaluation split; other backbones and splits may behave differently. The DP guarantee is per scan, but guarantees patient-level privacy only when each patient contributes one scan. The VLM and source cache are held by a trusted coordinating site and are not protected in this current version. Future work will investigate prototype selection and adaptive weighting so that new sites can improve predictions at other institutions, protection of the source cache, patient-level privacy guarantees, and differentially private encoder training.

## 6. CONCLUSION

SCAN enables hospitals to adapt shared brain MRI vision-language models without retraining or transferring patient scans. Local prototypes greatly improved classification at several unseen sites, and privacy-protected contributions preserved accuracy at existing sites, with near-chance membership inference. Cross-site gains were limited in our current tests, but SCAN establishes a practical framework to benchmark neurological disease classification across institutions while protecting patient privacy.

## 7. COMPLIANCE WITH ETHICAL STANDARDS

All datasets were collected with ethics board approval and anonymized and de-identified prior to analysis.

## 8. ACKNOWLEDGMENTS

This work was supported by the U.S. National Institutes of Health, under NIA grant U01 AG068057 and S10OD032285. We are grateful to the ENIGMA Stroke Working Group (https://enigma.ini.usc.edu/ongoing/enigma-stroke-recovery/) for providing the stroke dataset; the ENIGMA Epilepsy Working Group (https://enigma.ini.usc.edu/ongoing/enigma-epilepsy/) for providing the epilepsy dataset and the Department of Science and Technology, Govt. of India, grant nos. DST-SR/CSI/73/2011 (G); DST SR/CSI/70/2011 (G); and DST/CSRI/2017/249 (G) for the NIMHANS dataset.